# Deterministic Regime Switching and Feasibility Inversion in Dynamic Tensor Rematerialization

***An Empirical Study on the Reference DTR Simulator***

Mahesh Reddy Pagadala · Aston University · Preprint, September 2026
mahesh15116@gmail.com · Code and data: github.com/lonewolf15116/dtr-regime-switching

**Abstract.** We report fine-grained, deterministic instability in Dynamic Tensor Rematerialization (DTR), an online eviction policy for memory-constrained DNN training, measured on the reference DTR simulator (simrd) using public execution traces. On an LSTM trace, memory budgets differing by 0.10% of unconstrained peak memory select fast and slow execution regimes whose overheads differ by as much as 7.3×; the slow regime is driven by broadly repeated re-eviction of the same storages (evictions per storage rise from 1.33 to 8.27 while the set of distinct evicted storages is essentially unchanged: 5,233 vs 5,236, with the two sets overlapping at Jaccard 0.999). On a ResNet-32 trace, a fine budget sweep reveals a deterministic feasibility inversion: the run is feasible at ratio 0.101, infeasible (OOM) across 0.102-0.106, and feasible again from 0.107. We trace the immediate cause of the OOM to a fully pinned recursive rematerialization frontier that exceeds the budget after every evictable tensor has been evicted. Ablations using the DTR authors' own variants implicate the joint size-staleness scoring term in the observed LSTM instability. We argue these are at least two distinct budget-sensitive pathologies rather than one mechanism, and we separate what is demonstrated from what remains hypothesised. All results concern the reference simulator; reproduction in a production runtime is future work. Code, instrumentation, and raw results accompany this preprint.

## 1 Introduction

Training large neural networks requires caching intermediate activations from the forward pass for reuse in the backward pass. On memory-constrained hardware these activations may not fit at once, and tensor rematerialization - evicting some activations and recomputing them on demand - has become a standard remedy [3, 7]. Static methods such as Checkmate [4] solve an integer program over the full graph before training; dynamic methods such as DTR [6] decide evictions online using a local score, supporting dynamic graphs without ahead-of-time analysis.

DTR scores each resident tensor by compute cost, size, and staleness, evicting the lowest-scoring candidate when memory is exhausted; in the configuration studied here the score is compute(t) / (size(t) × staleness(t)). DTR's paper [6] proves an $\Omega(N/B)$ lower bound and reports strong empirical performance on a 0.05-step budget grid. It also already notes that DTR can suffer deep recursive rematerialization and, in its hardware prototype, out-of-memory conditions and hangs. We therefore do not claim to be first to observe that DTR can thrash or OOM. Our contribution is a fine-grained, deterministic characterisation of how these behaviours appear as a function of the memory budget, with an allocation-level mechanism for one and an ablation localising another.

To our knowledge (search scope: the DTR, Checkmate, Coop, MegTaiChi and T-Control lines and their immediate citations), the following combination has not previously been reported:

- Deterministic fine-grained overhead regime switching on LSTM: budgets separated by 0.001 in ratio select ~1.38× versus ~8.78-10.10× overhead, identical across three repeats.
- Deterministic non-monotone feasibility on ResNet-32: feasible, then a contiguous OOM band, then feasible again as the budget increases.
- An allocation-level mechanism for the ResNet OOM: a fully pinned recursive rematerialization frontier that exceeds the budget once the evictable pool is empty.
- Slow-regime structure on LSTM as broad re-eviction of nearly the entire working set, rather than a few pathological tensors dominating.
- Ablation evidence implicating the joint size-staleness term in the tested LSTM instability.

## 2 Background

simrd replays pre-recorded execution traces, capturing tensor operations, sizes and compute costs without a GPU. We use RuntimeV2EagerOptimized, the runtime supporting DTR's region-aware features. Each heuristic returns a score and the lowest-scoring storage is evicted. The variants used are DTR = compute/(size × staleness); AbESize = compute/size (staleness removed); AbEStale = compute/staleness (size removed); AbE = compute (both removed); the three ablations are from the DTR authors' ablation.py. Overhead is (model compute + recomputation) / model compute; 1.0 means no recomputation. Budget ratio is the memory cap divided by unconstrained peak memory. A run that exhausts memory raises MemoryError, reported as OOM; this is distinct from exceeding a recomputation ceiling.

## 3 Experiments

### 3.1 Experimental setup and reproducibility

Simulator: the DTR public artefact (github.com/uwsampl/dtr-prototype), simrd commit eff53cc4804cc7d6246a6e5086861ce2b846f62b. Traces: the artefact's own recordings for LSTM (lstm-128-…-default.log) and ResNet-32 (resnet32-56-…-default.log). Runtime class RuntimeV2EagerOptimized; heuristic classes DTR, AbESize, AbEStale, AbE. Budget in bytes is int(peak × ratio). Each point is run as a fresh runtime; the ResNet OOM band was additionally reproduced in independent OS processes. Three termination classes are distinguished programmatically: OOM (MemoryError from the runtime's _free when the evictable pool is empty and the allocation still does not fit), recomputation-limit termination (a separate exception, not used for the OOM results), and Python RecursionError (stack exhaustion during deep rematerialization). For the OOM and frontier probes the recomputation limit is set to infinity so an OOM is never pre-empted by an overhead cap; deep recursion is enabled with a raised recursion limit and a large thread stack. These changes widen limits and add read-only logging; they do not alter eviction decisions. Results reproduced identically on Python 3.11 (Linux) and Python 3.13 (Windows). Scripts, instrumentation, exact commands and raw per-point results accompany this preprint.

### 3.2 Models and baselines

LSTM has unconstrained peak 5,595 MB and a single-op bottleneck of 69.4 MB (1.2% of peak); ResNet-32 has peak 10,061 MB and bottleneck 539.5 MB (5.4% of peak). LSTM is recurrent; ResNet-32 is feedforward with residual connections. Low-budget behaviour is characterised with linear scans rather than assuming a curve shape. The evaluated simulator configuration is deterministic for the tested traces and settings, so we run three repeats per point and report whether they agree.

### 3.3 LSTM: deterministic overhead regime switching

Selected points from a fine 0.001-step sweep on LSTM show both a fast and a slow overhead regime, including switches between adjacent budgets (Table 1); all three repeats at each budget were identical to four decimal places (no variation observed).

| **Budget ratio** | **Budget (MB)** | **DTR overhead** | **Regime** |
|---|---|---|---|
| 0.268 | 1,500 | 1.380× | fast |
| 0.266 | 1,488 | 10.100× | SLOW |
| 0.265 | 1,483 | 1.383× | fast |
| 0.264 | 1,477 | 8.779× | SLOW |
| 0.262 | 1,466 | 1.387× | fast |

*Table 1. LSTM overhead at selected fine-grid budgets (three repeats each, identical at every reported point). Rows are selected points from the 0.001-step sweep; adjacent-budget transitions include 0.266→0.265 and 0.265→0.264. The 0.266→0.265 step is 0.10% of unconstrained peak memory (~0.38% of the constrained budget).*

### 3.4 LSTM slow-regime mechanism: broad working-set re-eviction

Comparing the fast budget (0.265) and the adjacent slow budget (0.264) at the storage level (Table 2), the distinct-storage count is essentially unchanged (5,233 vs 5,236) while total evictions rise 6.2×. Comparing the two evicted-storage sets by identity confirms this is repeated re-eviction of the same storages, not expansion to new tensors: all 5,233 storages evicted in the fast regime are also evicted in the slow regime, which adds only 3 further storages (set overlap Jaccard 0.999). The most re-evicted storage rises from 5 to 24 evictions, and the top-10 storages account for under 1% of evictions in both regimes - broad recycling of nearly the whole working set, not a small hot set. Inspection of the ordered eviction trace revealed repeatedly occurring short subsequences, consistent with periodic working-set recycling; we report this qualitatively and do not quantify a period. Collateral (dependency-driven) rematerialization is 0.0% in both regimes by the simulator's measure, providing no evidence for deep rebuild chains in this LSTM case.

| Metric | FAST (0.265) | SLOW (0.264) |
|---|---|---|
| Overhead | 1.383× | 8.779× |
| Total evictions | 6,958 | 43,292 |
| Distinct storages evicted | 5,233 | 5,236 |
| Evictions per storage | 1.33 | 8.27 |
| Most re-evicted storage | 5× | 24× |
| Top-10 storages' share | 0.7% | 0.5% |
| Collateral remat. | 0.0% | 0.0% |

*Table 2. Storage-level comparison of the LSTM fast and slow regimes. The two evicted-storage sets overlap at Jaccard 0.999 (fast set fully contained in the slow set), confirming re-eviction of the same storages rather than expansion to new tensors.*

### 3.5 Ablations: the joint size-staleness term

At the same five LSTM budgets, only full DTR alternates; all three ablations stay within a narrow overhead band (Table 3), each deterministic across repeats. Within this window and these variants, removing either size or staleness eliminates the switching, implicating their joint presence in the denominator. At DTR's two slow points (0.266, 0.264), AbEStale has 4.8× and 6.6× lower overhead respectively. This is an association within the tested window; it does not by itself establish the score-level causal path (Section 6).

| Heuristic | Formula | Switches? | Range (5 budgets) |
|---|---|---|---|
| DTR | compute/(size × staleness) | Yes | 1.38-10.10× |
| AbESize | compute/size | No | 5.65-6.15× |
| AbEStale | compute/staleness | No | 1.33-2.09× |
| AbE | compute | No | 5.51-6.29× |

*Table 3. DTR vs the three DTR-paper ablations across the five LSTM budgets.*

### 3.6 ResNet-32: deterministic feasibility inversion

A 0.001-step sweep from 0.100 to 0.150 (three repeats, all deterministic) shows two things the coarse grid hides. First, overhead is piecewise-constant and non-monotone: it rises from 2.077× at 0.107 to 3.212× at 0.111 and 3.330× at 0.116 as the budget increases, then falls back (Figure 1). Second, feasibility is non-monotone: the run is feasible at 0.100-0.101, OOM across the contiguous band 0.102-0.106, and feasible again from 0.107 onward (Table 4). A larger budget thus makes a run fail that a smaller budget completes. The band was reproduced in independent single-point processes and with the recomputation limit disabled.

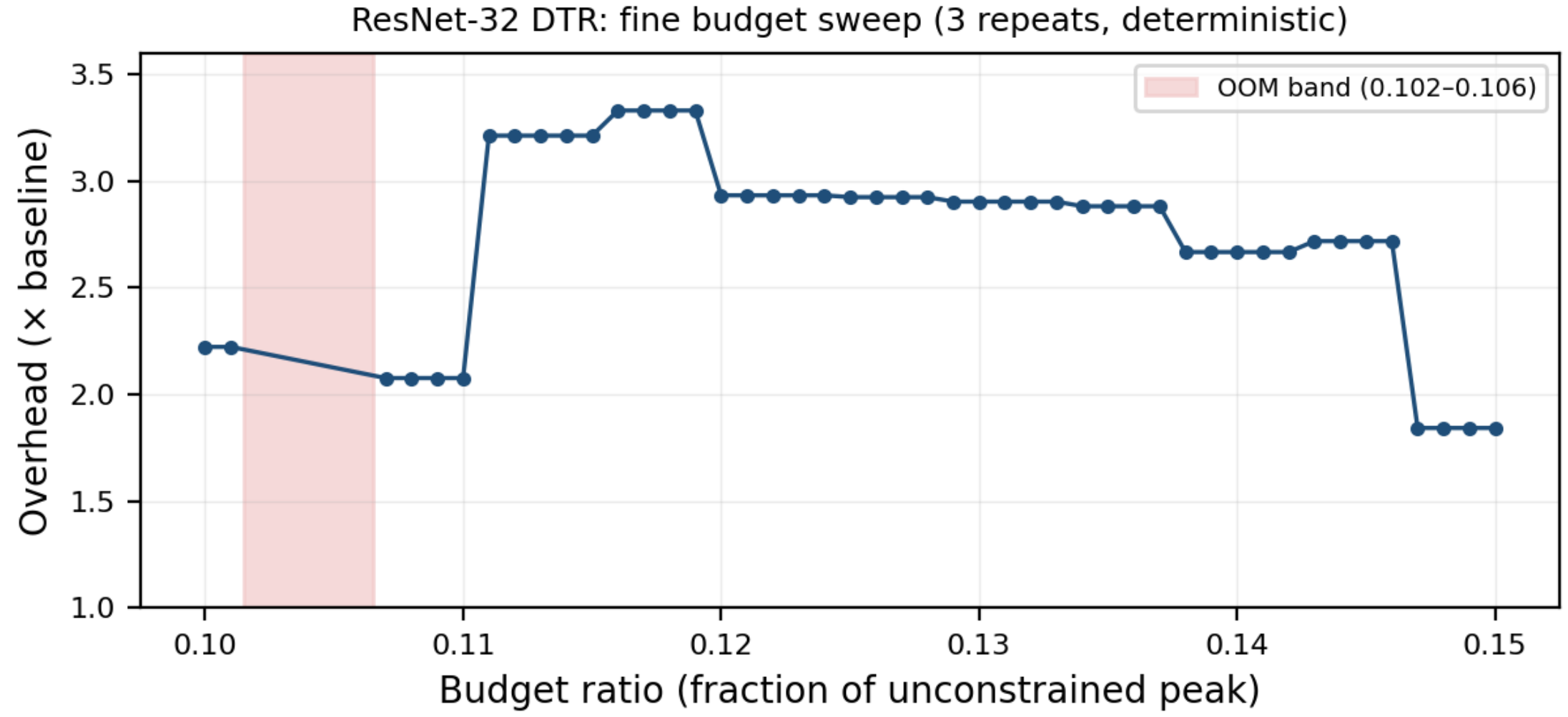

*Figure 1. ResNet-32 DTR overhead across the full 0.100-0.150 sweep (0.001 steps, three deterministic repeats). Shaded: the 0.102-0.106 OOM band. Overhead is non-monotone and piecewise-constant.*

Table 4 gives the complete sweep as contiguous bands (every 0.001 point is covered; per-point raw values accompany the preprint).

| Budget band | Outcome / overhead |
|---|---|
| 0.100-0.101 | 2.223× (feasible) |
| 0.102-0.106 | **OOM** |
| 0.107-0.110 | 2.077× |
| 0.111-0.115 | 3.212× |
| 0.116-0.119 | 3.330× |
| 0.120-0.124 | 2.933× |
| 0.125-0.128 | 2.924× |
| 0.129-0.133 | 2.903× |
| 0.134-0.137 | 2.881× |
| 0.138-0.142 | 2.667× |
| 0.143-0.146 | 2.718× |
| 0.147-0.150 | 1.843× |

*Table 4. Complete ResNet-32 DTR fine sweep, 0.100-0.150 (contiguous bands; all points deterministic over three repeats). Feasible → OOM → feasible as the budget increases is the central observation.*

### 3.7 ResNet-32 OOM mechanism: a fully pinned recursive frontier

At ratio 0.104 the run OOMs while materialising tensor x7977, which requests 179.8 MB. At the failure the evictable pool is empty and 889.8 MB is resident, all of it locked across 231 distinct storages formed by a recursive rematerialization chain of depth 118; 889.8 + 179.8 = 1069.6 MB exceeds the 1046.4 MB budget by 23.2 MB. Because the pool is empty, the failure is not caused by overlooking an evictable victim at the failing allocation - the evictable storage is exhausted and the remainder is pinned.

Comparing the peak pinned frontier across three budgets (Table 5, Figure 2) shows the frontier is budget-dependent. The run-level maximum frontier depth is itself non-monotone (8, 118, 30), though the 0.101 maximum occurs at a different tensor (x8106) than the 0.104 and 0.107 maxima (x7977), so this is a run-level, not a tensor-matched, comparison. The clean same-tensor comparison is x7977: at 0.104 it rebuilds at depth 118 with 889.8 MB pinned and OOMs; at the larger budget 0.107 it rebuilds at depth 30 with 575.1 MB pinned and completes. The divergent trajectories imply that different prior eviction histories precede the observed frontiers; the responsible score and victim-order transitions remain untraced.

| Measurement | 0.101 | 0.104 | 0.107 |
|---|---|---|---|
| Max pinned bytes (MB) | 754.9 | 889.8 | 575.1 |
| Max distinct pinned storages | 227 | 231 | 226 |
| Recursion depth at peak | 8 | 118 | 30 |
| Tensor at peak frontier | x8106 | x7977 | x7977 |
| Requested at peak (MB) | 179.8 | 179.8 | 179.8 |
| Pinned + requested (MB) | 934.7 | 1069.6 | 754.9 |
| Budget (MB) | 1016.2 | 1046.4 | 1076.5 |
| Outcome | feasible | **OOM** | feasible |

*Table 5. Peak pinned rematerialization frontier at three budgets (recomputation limit disabled; storages de-duplicated by identity). The same-tensor x7977 rows are the directly comparable evidence.*

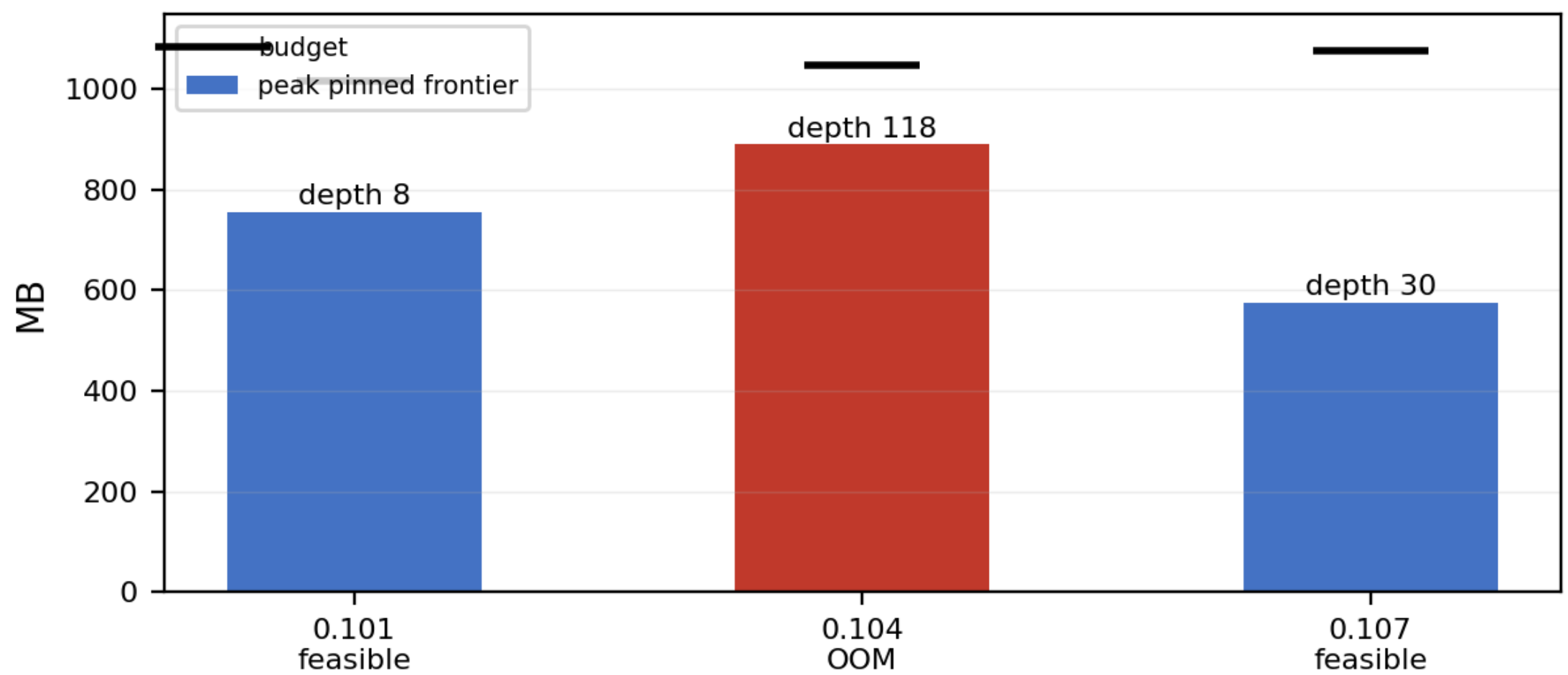

*Figure 2. Peak pinned frontier vs budget at ratios 0.101/0.104/0.107, annotated with recursion depth. At 0.104 the pinned frontier plus the pending 179.8 MB allocation exceeds the budget, after the evictable pool is empty, producing the OOM.*

### 3.8 Interpretation: at least two distinct pathologies

The LSTM slow regime (broad working-set re-eviction, ~43k evictions, 0% collateral) and the ResNet OOM (a deep, fully pinned recursive frontier) are structurally different failure modes. They may share an upstream cause - budget-dependent victim selection - but we do not demonstrate that they are the same mechanism, and we do not force them into one. The RecursionError observed on LSTM at very low budgets is a plausible third manifestation of frontier depth, but stack exhaustion is partly an implementation property and we treat it as suggestive only.

## 4 Relation to Prior Work

DTR [6] proves an $\Omega(N/B)$ lower bound, reports strong results on a 0.05-step grid, and explicitly notes deep recursive rematerialization and prototype OOMs/hangs - so thrashing and OOM in DTR are known. Our results add the deterministic fine-grained structure between grid points: overhead regime switching, a feasibility inversion, an allocation-level frontier mechanism, and an ablation localisation. T-Control [8] targets a different failure mode (eviction of structurally important tensors) and reports coarser budget increments that would not sample the narrow transitions here; its score also retains size and staleness. Coop [9] and MegTaiChi [5] address fragmentation and allocation. Belady's anomaly [1] - more cache can raise FIFO page faults - is a useful structural analogy for non-monotone behaviour under a non-stack policy (stack-based policies such as LRU cannot exhibit it [2]), but the objects differ (recomputation and lock lifetimes vs paging), so we claim analogy, not equivalence. A systematic review is required before any literature-wide novelty claim.

## 5 Implications

Evaluation grids. Coarse budget grids can miss narrow transitions and yield comparisons that depend on the sampled points; fine sweeps with repeats are needed near high-pressure regions, and the resolution should be reported.

Feasibility-boundary search. Because feasibility is empirically non-monotone in the observed ResNet interval, bisection is not a reliable method for identifying the global minimum feasible budget for this configuration. We make no universal claim about all rematerialization systems; linear scans are a safe default for mapping the local boundary.

Heuristic design. DTR combines a memory-reclamation signal (size) and recomputation-risk signals (compute, staleness) in one multiplicative score. The LSTM ablation implicates their joint term, while ResNet shows size cannot simply be dropped without harming reclamation under tight pressure. A candidate direction is to separate the objectives - enforce a minimum memory-release criterion, then rank eligible candidates by recomputation risk - evaluated against full DTR, its ablations, and offline or clairvoyant references. This is a hypothesis, not a validated result.

## 6 Scope of Claims

We separate the claims by evidential status.

- **Demonstrated (this simulator, these traces):** deterministic overhead regime switching on LSTM; deterministic feasibility inversion on ResNet-32; the allocation-level OOM mechanism (pinned frontier exceeds budget with the evictable pool empty); broad working-set re-eviction in the LSTM slow regime; identical repeats.
- **Supported but incomplete:** that earlier victim history produces the divergent ResNet frontier (shown at the allocation and frontier level; not traced to individual score comparisons).
- **Hypothesised:** that the LSTM and ResNet pathologies share a single upstream size-staleness or victim-order mechanism; that a decoupled policy removes the instability without harming feasibility.
- **Unknown:** whether any of this reproduces in the production PyTorch/GPU DTR runtime under real allocators and asynchronous execution. This is the primary open risk and the priority for follow-up.

## 7 Conclusion

On the reference DTR simulator, small memory-budget changes deterministically switch LSTM between fast and slow overhead regimes and drive ResNet-32 through a feasible→OOM→feasible inversion. The LSTM slow regime is broad re-eviction of nearly the whole working set; the ResNet OOM is a fully pinned recursive rematerialization frontier that exceeds the budget once evictable storage is exhausted. Ablations implicate the joint size-staleness term for the LSTM instability. We present these as at least two distinct budget-sensitive pathologies with a shared-cause hypothesis, and we mark the score-level mechanism and runtime reproduction as open. All results are reproducible from the public simulator and traces via the accompanying package.